\ifdefined\pdfobjcompresslevel
\fi

\documentclass[11pt]{article}

\usepackage[preprint]{acl}

\usepackage{times}
\usepackage{latexsym}
\usepackage[T1]{fontenc}

\usepackage[utf8]{inputenc}

\usepackage{microtype}

\usepackage{inconsolata}

\usepackage{graphicx}
\usepackage{booktabs} 
\usepackage{array}
\usepackage{amsmath}
\usepackage{enumitem}
\usepackage{tablefootnote}
\usepackage{multirow}
\usepackage{makecell}
\usepackage{adjustbox}
\usepackage[table]{xcolor}
\usepackage{booktabs,array,xcolor,colortbl,makecell}
\usepackage[most]{tcolorbox}
\usepackage{hyperref}

\newtcolorbox{observationbox}{
  colback=gray!8,
  colframe=gray!35,
  boxrule=0.4pt,
  arc=3pt,
  left=8pt,
  right=8pt,
  top=6pt,
  bottom=6pt,
  width=\linewidth,
  enhanced
}

\newenvironment{observationbox*}
{
\begin{figure*}[t]
\centering
\begin{tcolorbox}[
  colback=gray!8,
  colframe=gray!35,
  boxrule=0.4pt,
  arc=3pt,
  left=8pt,
  right=8pt,
  top=6pt,
  bottom=6pt,
  width=0.96\textwidth,
  enhanced
]
}
{
\end{tcolorbox}
\end{figure*}
}

\usepackage{tikz,xcolor}

\definecolor{capblue}{HTML}{59C7F2}

\newcommand{\capstrong}{%
  \tikz[baseline=-0.6ex]\draw[black, fill=capblue] (0,0) circle (0.9ex);%
}

\newcommand{\capweak}{%
  \tikz[baseline=-0.6ex]\draw[black, fill=white] (0,0) circle (0.9ex);%
}

\newcommand{\caplimited}{%
  \tikz[baseline=-0.6ex]{
    \begin{scope}
      \clip (0,0) circle (0.9ex);
      \fill[capblue] (-0.9ex,-0.9ex) rectangle (0,0.9ex);
    \end{scope}
    \draw[black] (0,0) circle (0.9ex);
  }%
}

\newcommand{\capmoderate}{%
  \tikz[baseline=-0.6ex]{
    \begin{scope}
      \clip (0,0) circle (0.9ex);
      \fill[capblue] (-0.9ex,-0.9ex) rectangle (0.0ex,0.9ex);
    \end{scope}
    \draw[black] (0,0) circle (0.9ex);
  }%
}

\newtcbox{\pillstrong}{on line,
  boxrule=0pt, arc=10pt, left=8pt, right=8pt, top=3pt, bottom=3pt,
  colback=stronggreen, coltext=white, boxsep=1pt}
\newtcbox{\pillweak}{on line,
  boxrule=0pt, arc=10pt, left=8pt, right=8pt, top=3pt, bottom=3pt,
  colback=weakyellow, coltext=black, boxsep=1pt}
\newtcbox{\pillmoderate}{on line,
  boxrule=0pt, arc=10pt, left=8pt, right=8pt, top=3pt, bottom=3pt,
  colback=moderateblue, coltext=black, boxsep=1pt}
\newtcbox{\pilllimited}{on line,
  boxrule=0pt, arc=10pt, left=8pt, right=8pt, top=3pt, bottom=3pt,
  colback=limitedgray, coltext=black, boxsep=1pt}

\usepackage{soul}
\colorlet{soulblue}{blue!20}

\newif\ifhidecomment
\ifhidecomment
\newcommand{\yz}[1]{}
\newcommand{\tx}[1]{}
\newcommand{\cm}[1]{}
\newcommand{\dm}[1]{}
\newcommand{\ds}[1]{}
\newcommand{\yw}[1]{}
\newcommand{\red}[1]{#1}
\else
\newcommand{\yz}[1]{\sethlcolor{soulblue}\hl{[Evelyn: #1]}}
\newcommand{\yw}[1]{\sethlcolor{pink!60}\hl{[Yvonne: #1]}}
\newcommand{\dm}[1]{\sethlcolor{red!60}\hl{[Daniel: #1]}}
\newcommand{\ds}[1]
{\sethlcolor{yellow!60}\hl{[Dimitris: #1]}}
\newcommand{\tx}[1]
{\sethlcolor{pink}\hl{[Tong:#1]}}
\newcommand{\cm}[1]{\sethlcolor{orange!60}\hl{[Cecilia:#1]}}
\newcommand{\red}[1]{#1}
\fi

\newcommand{\grayrow}{\rowcolor[gray]{.9}}
\definecolor{deemph}{gray}{0.55}
\newcolumntype{g}{>{\columncolor[gray]{0.9}}c}
\newcolumntype{R}[1]{>{\raggedleft\arraybackslash}p{#1}}

\title{\protect\includegraphics[height=1.1em]{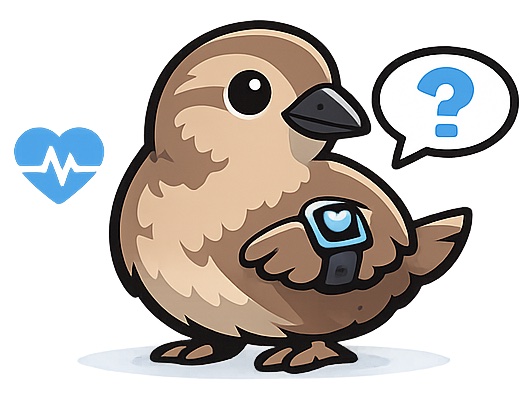}WEQA: \underline{W}earable {{h\underline{E}}}alth \underline{Q}uestion \underline{A}nswering with \\
Query-Adaptive Agentic Reasoning}

\author{
 \textbf{Yuwei Zhang\textsuperscript{1}},
 \textbf{Tong Xia\textsuperscript{1,2}},
 \textbf{Bianca Emmerich\textsuperscript{3}},
 \textbf{Yu Yvonne Wu\textsuperscript{4}},
\\
 \textbf{Dimitris Spathis\textsuperscript{1,5}},
 \textbf{Xin Liu\textsuperscript{5}},
 \textbf{Daniel McDuff\textsuperscript{5}},
 \textbf{Cecilia Mascolo\textsuperscript{1}}
\\
 \textsuperscript{1}University of Cambridge,
 \textsuperscript{2}Tsinghua University,
 \textsuperscript{3}University College London, \\
 \textsuperscript{4}Dartmouth College,
 \textsuperscript{5}Google Research
\\
 \small{
   \texttt{yz798@cl.cam.ac.uk}
 }
}

\begin{document}
\maketitle

\newenvironment{Itemize}{
 \begin{itemize}[leftmargin=*]
     \setlength{\itemsep}{0pt}
     \setlength{\topsep}{0pt}
     \setlength{\partopsep}{0pt}
     \setlength{\parskip}{0.5pt}}
 {\end{itemize}}

 \newenvironment{Enumerate}{
 \begin{enumerate}[leftmargin=*]
     \setlength{\itemsep}{0pt}
     \setlength{\topsep}{0pt}
     \setlength{\partopsep}{0pt}
     \setlength{\parskip}{1pt}}
 {\end{enumerate}}
 
\begin{abstract}
Language models are remarkably capable at medical question answering, in some cases surpassing the accuracy of general physicians. However, answering questions about wearable health data remains challenging and understudied, as these ubiquitous sensors produce continuous, high-dimensional, and longitudinal data, which is non-trivial to align with text-centric distributions in LLM pretraining. The diversity of sensor modalities and user intents cannot be effectively handled by a fixed reasoning workflow or a single pretrained foundation model.
To address these challenges, we propose \textbf{WEQA}, a query-adaptive 
agent framework that unifies LLM reasoning with specialized wearable analytical and modeling tools.
An LLM controller is employed
to synthesize execution plans and 
\textit{dynamically route}
each query to the appropriate combination of sensor analysis and pretrained models, and perform grounded response auditing with external knowledge.
We also curate a benchmark spanning four open wearable datasets comprising analytic and predictive tasks in three different health domains. Experiments show that our framework is 24\% more accurate than
LLM and agentic baselines, and a blinded study with 12 medical experts and 8 users shows substantial gains in usefulness and clinical soundness.
\end{abstract}

\section{Introduction}

\begin{figure}[t]
  \includegraphics[width=\columnwidth]{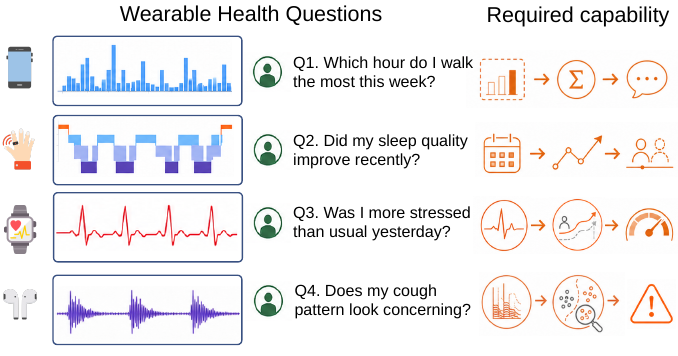}
  \caption{\textbf{Examples of heterogeneous wearable health questions and the computational requirements.} The right panel denotes representative computation steps, such as  temporal analysis and model prediction. 
  }
  \label{fig:task_examples}
\end{figure}

Smartwatches and other wearable devices are widely adopted and have transformed passive and continuous monitoring of activity and health~\cite{cosentino2024towards, xu2025lsm}.  Beyond tracking wellness indicators such as sleep, activity, and stress, wearable sensors support early detection of conditions including depression, hypertension, and respiratory diseases~\cite{zhang2024towards, xu2025lsm} and potentially life saving interventions~\cite{shah2025automated}.
 Meanwhile, large language models (LLMs) are making ubiquitous conversational health assistance increasingly plausible~\cite{openai_chatgpt_health_2026}. 
 Language models can be used to interpret free-form questions, provide explanations and reasoning, and draw on broad medical knowledge acquired during pretraining. 
 Together, these advances present a compelling vision of \textbf{wearable health question answering} (Fig.~\ref{fig:task_examples}), where users ask natural-language questions about their sensor data and receive personalized, clinically meaningful responses~\cite{heydari2025anatomy}.

Nevertheless, reasoning over physiological and behavioral time series is different from text reasoning~\cite{langer2025opentslm}. 
Wearable signals are continuous, high-dimensional, and inherently temporal. Their interpretation depends on subtle waveform morphology, long-range temporal dynamics, and cross-sensor interactions (Fig.~\ref{fig:task_examples}). 
\textit{E.g.}, sleep improvement questions require reasoning over longitudinal behavioral patterns, while blood-pressure-related questions depend on subtle photoplethysmogram morphology. Identical aggregate statistics may correspond to clinically different physiological states depending on temporal structure and individual variations. Hence, \textit{wearable health reasoning requires native understanding of physiological signals rather than textual abstractions}.

Another key challenge is the diversity of wearable health questions and the heterogeneous reasoning requirements. As  Fig.~\ref{fig:task_examples} shows, queries span descriptive analytics, longitudinal trend monitoring, personalized mental health assessment, and predictive screening from raw signals. These tasks require fundamentally different forms of computation, from lightweight statistical aggregation (Q1) to long-horizon temporal reasoning (Q2), model prediction from raw physiological signals (Q3\&4), and personalized evidence synthesis (Q3). Consequently, \textit{wearable health question answering cannot be effectively supported through a single fixed reasoning workflow or prediction model, but requires adaptive computational pathways tailored to the query intent and physiological context.}


\begin{table}[t]
\centering
\tiny
\setlength{\tabcolsep}{2.5pt}
\renewcommand{\arraystretch}{1.1}
\begin{tabular}{R{2.5cm} c c c g}
\toprule
 & \makecell{Text-only \\LLMs}
 & \makecell{LLM \\ Coding Agents}
 & \makecell{Wearable\\Foundation Models}
 & \makecell{\textbf{WEQA}} \\
\midrule
 \textbf{Language Comprehension}  & \capstrong
& \capstrong
& \capweak
& \capstrong \\
\textbf{Statistical Analysis} & \capweak
& \capstrong
& \caplimited
& \capstrong \\

\textbf{Temporal Reasoning} & \capweak
& \capmoderate
& \capmoderate
& \capstrong \\

\textbf{Predictive Inference} & \capweak
& \capweak
& \capstrong
& \capstrong \\

\textbf{Adaptive Routing} & \capweak
& \capmoderate
& \capweak
& \capstrong \\

\textbf{Personalization} & \capweak
& \capweak
& \capmoderate
& \capstrong \\
\textbf{Safety \& Grounding} & \capweak
& \capmoderate
& \capweak
& \capstrong \\
\bottomrule
\multicolumn{5}{r}{ \capstrong\ Strong 
\quad \caplimited\ Limited \quad \capweak\ Weak}
\end{tabular}
\caption{\textbf{Capability comparison across wearable health QA paradigms.} Symbols indicate capability levels of strong, limited, and weak.}
\vspace{-10pt}
\label{tab:wearable_qa_comparison}
\end{table}

To this end, we propose \textbf{WEQA} (\textbf{W}earable h\textbf{E}alth \textbf{Q}uestion \textbf{A}nswering), a query-adaptive agent framework that combines LLM-based reasoning with \red{specialized wearable modeling}. 
To overcome the misalignment between wearable sensor data and text-centric LLM pretraining,
we develop a \textit{wearable-health toolset} that enables sensor-native understanding directly from raw physiological signals, including temporal, cross-sensor analysis and specialized predictive modeling, \red{such as machine learning and foundational models}. This leverages fine-grained physiological dynamics structure that are often lost in textual summaries, and powerful models optimized for wearable health data.

To support diverse wearable health question answering,  we introduce a \textit{Query-Aware Planning} mechanism that uses the LLM as a controller to synthesize execution plans conditioned on the user query and available sensor context. Instead of relying on a fixed reasoning pipeline or a single pretrained model~\cite{li2025sensorllm, yu2025sensorchat}, the agent \textit{dynamically} selects the computation pathway from the toolset for each question. 
We also design a \textit{Grounded Response Auditing} phase that verifies whether responses are supported by sensor evidence and external medical knowledge, while calibrating risk-sensitive language. Together, WEQA enables a unified framework for wearable health question answering across heterogeneous tasks, sensing modalities, and personalization settings.

To support systematic evaluation, we curate a 
benchmark spanning four open wearable datasets across cardiovascular, respiratory and mental health, covering both analytical and predictive tasks. We organize the benchmark into four task families: \textit{Short-Horizon Analytical QA}, \textit{Long-Horizon Analytical QA}, \textit{Predictive Reasoning QA}, and \textit{Open-Ended Insight QA}. These tasks require different combinations of \red{analysis, prediction, and clinical calibration}.
Across tasks, our framework substantially outperforms LLM-only and existing agentic baselines by at least 24\%, demonstrating the value of adaptively combining sensor-native modeling with LLM reasoning.  
We further conduct expert evaluation showing that our framework produces more useful and clinically sound responses, moving beyond benchmark accuracy toward realistic health-assistance quality.
Ablation studies show that the gains are not from the wearable toolset alone: removing the query-adaptive workflow leads to lower accuracy and consistent degradation across all human-rated dimensions.

Our work makes four main contributions. \textbf{(1) Challenges}: We identify two core challenges in wearable health question answering: bridging complex physiological signals with text-centric LLMs, and the heterogeneity of health queries that demands adaptive reasoning workflows. \textbf{(2) WEQA Framework}: We introduce a novel query-adaptive agent architecture that dynamically routes between sensor reasoning, adaptive predictive inference, and grounded response auditing, supported by a reusable wearable-health toolset. \textbf{(3) Benchmark}: We introduce a new benchmark for evaluating wearable health question answering across diverse analytical and predictive tasks. \textbf{(4) Evaluation}: We demonstrate through automated and human evaluation that WEQA provides a stronger foundation for personalized, grounded, and safety-aware wearable health assistance over LLM and agent baselines.

\section{Related work}

\begin{figure*}[t]
\vspace{-10pt}
  \includegraphics[width=\linewidth]{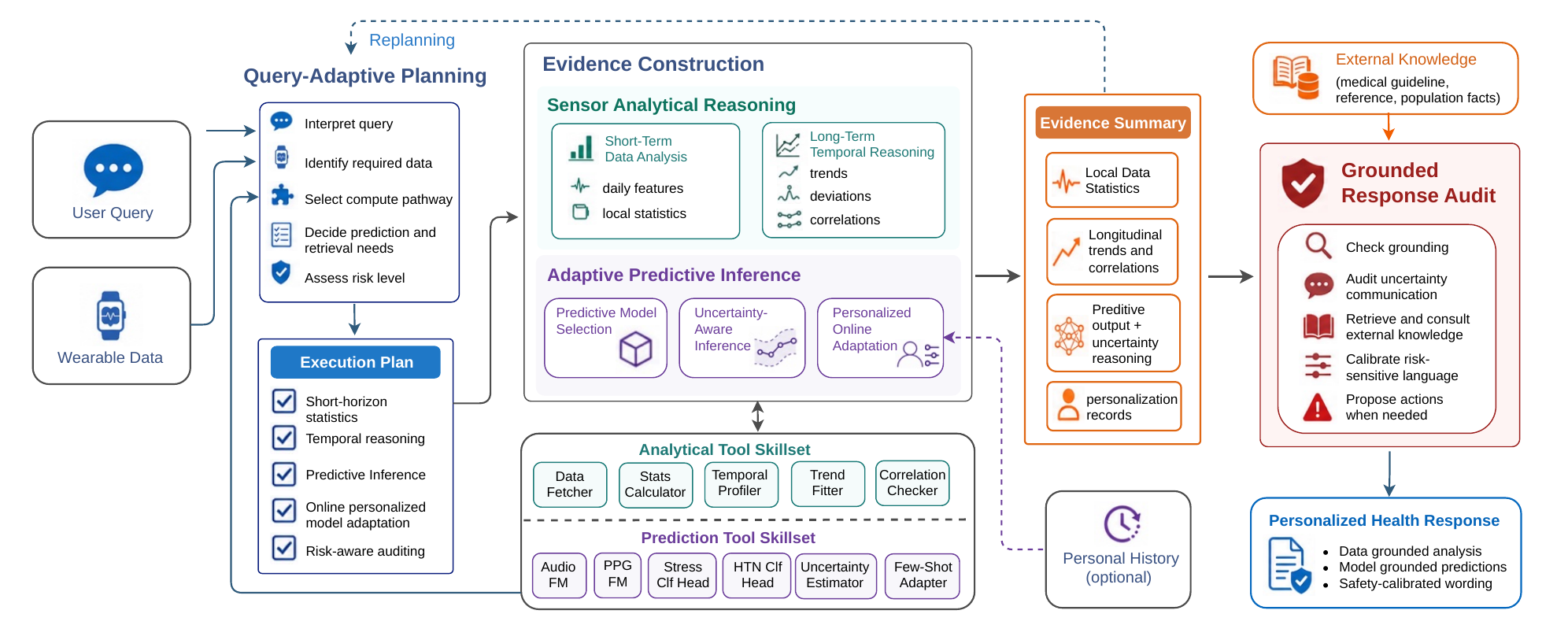}
\caption{\textbf{WEQA query-adaptive agent framework for wearable health question answering}. Given a user query, wearable data, and optional personal history, the agent plans a query-specific workflow, gathers evidence through sensor analysis and adaptive prediction, and performs grounded safety auditing producing a safety-aware response. 
}
  \label{fig:workflow}
  \vspace{-5pt}
\end{figure*}

\textbf{Wearable Health LLMs and Agents.}
A common approach for wearable health LLMs is to summarize sensor data as text and let an LLM reason over the prompt~\cite{liu2023large,kim2024health, cosentino2024towards}.
While simple and flexible, text-based representations discard fine-grained \textit{temporal structure, signal morphology, and cross-sensor dynamics} that are critical for physiological interpretation, limiting performance on more complex wearable-health tasks~\cite{zhang2025sensorlm}. Recent agentic LLM systems extend this paradigm through planning, tool use, and multi-step reasoning~\cite{merrill2026transforming, heydari2025anatomy}, enabling grounded responses beyond one-shot prompting.
However, existing frameworks still largely rely on pre-aggregated coarse features as input and are limited to statistical analysis. In contrast, real-world wearable health question answering requires sensor-native understanding together with adaptive reasoning workflows that can generalize across diverse tasks, sensing modalities, temporal scopes, and personalization settings.

\vspace{0.2em}
\noindent \textbf{Wearable Health Foundation Models.}
Large-scale pretraining on wearable data has enabled transferable sensor-native representations and improved performance across diverse health tasks, including activity recognition, cardiovascular monitoring, respiratory screening, and mental health assessment~\cite{zhang2024towards,pillai2024papagei,xu2025lsm,narayanswamy2026generalintelligenceinterfacewearable}. 
Recent sensor-language models further explore aligning wearable signals with language interfaces for specific domains such as activity and sleep understanding~\cite{li2025sensorllm, zhang2025sensorlm, xu2026sleeplm}. Our work addresses a different problem: \textit{general wearable health question answering}. 
Existing wearable foundation models are typically designed for predefined modalities and fixed target domains, making adaptation to broader health queries computationally expensive and difficult to scale.
Rather than training a single universal model, WEQA introduces a unified agent framework that combines LLM reasoning with specialized sensor-native analytical tools and pretrained models. Conditioned on the user query and sensor context, the framework dynamically composes the computation pathway required for each question, framing wearable health question answering as an \textit{adaptive reasoning problem over heterogeneous physiological evidence rather than a fixed prediction task}.

\vspace{0.3em}
\noindent \textbf{Adaptive Agentic Frameworks.}
Recent work on adaptive agentic systems explores automated workflow design and modular agent optimization~\cite{hu2025automated, shang2025agentsquare, zhang2025aflow}.  
These approaches optimize workflows at the dataset or benchmark level through iterative search over training and validation sets. In contrast, wearable health agents need to adapt computation dynamically {\em at inference time based on the user query, sensor context, and personalization needs}. 
Our work addresses this setting through a training-free, query-adaptive framework.

\section{Method}

We propose \textbf{WEQA}, a query-adaptive agent framework for \textbf{W}earable h\textbf{E}alth \textbf{Q}uestion \textbf{A}nswering. 
The framework consists of three stages (Fig.~\ref{fig:workflow}): query-aware planning, evidence construction, and grounded response auditing.

\begin{figure*}[t]
\centering
\vspace{-5pt}
  \includegraphics[width=0.9\textwidth]{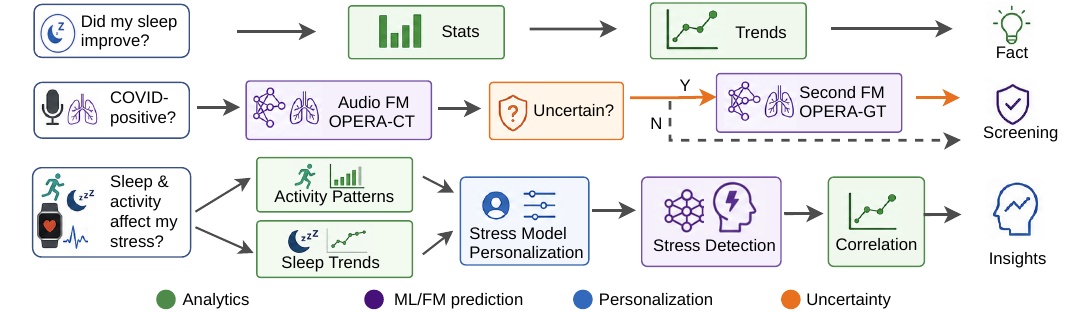}
  \caption{Examples of query-adaptive reasoning pathways in WEQA for heterogeneous wearable health questions.}
  \vspace{-10pt}
  \label{fig:pathways}
\end{figure*}
\vspace{0.3em}

\noindent \textbf{Problem Formulation.} Given a user query $q$, wearable sensor data $X$, and optional personal history $H$, the goal is to generate a  health response $y$. 

As Figure.~\ref{fig:workflow} shows, WEQA operates in three stages: (1) \textit{query-aware planning}, which maps the query to an execution plan $a = \pi(q, X, H)$; (2) \textit{evidence construction}, which gathers evidence $Z = \Phi(X, H, q, a)$ through analytical tools and predictive models; and (3) \textit{grounded response auditing}, which generates the final response $y = g(q, Z, H)$ after verifying evidential grounding and enforcing safety considerations. Here, $\pi$ denotes the planner, $\Phi$ the evidence construction process, and $g$ the auditing module. The adaptive design enables WEQA to tailor computation to diverse health queries while maintaining explicit sensor evidence.

\subsection{Query-Aware Planning}

The first stage converts the user query into a structured execution plan. The planner is implemented as a prompted LLM controller. Given the query $q$, sensor metadata $m$, available analytical tools and predictive models skillset $\mathcal{S}$, and retrieved examples $e$ from a history bank, the controller produces a structured plan $a = \pi(q, m, e, S)$. Motivated by the heterogeneous nature of wearable health queries, the controller first
identifies the query objective, required data modalities, temporal scope, reasoning type, personalization needs, and response risk level.
It then plans the analysis steps, required tools or models and expected evidence.




Planning is also not fixed. After each execution step $t$, the controller may revise the plan based on the accumulated evidence
$s_t = \{q, a_t, Z_t\}, $
where $a_t$ is the current plan and $Z_t$ is the accumulated evidence. This enables the agent to handle heterogeneous queries and recover from execution errors or missing data. Simple descriptive questions are routed to lightweight analysis, while higher-risk or predictive queries (\textit{e.g.}, hypertension classification) invoke specialized models, uncertainty-aware verification, and stricter response control.

\subsection{Evidence Construction}

The second stage executes the plan and constructs a sensor-based evidence summary $Z$. We distinguish two complementary execution pathways: sensor analytical reasoning and adaptive predictive inference, both orchestrated by the LLM controller. WEQA is equipped with wearable-health tools spanning statistical analysis, temporal reasoning, stress inference, respiratory screening, and cardiovascular prediction. 
More details in Appendix~\ref{sec:app:implement}.

\paragraph{Sensor Analytical Reasoning.}
For queries that can be answered directly from wearable observations, the agent executes code and invokes analytical tools over the relevant sensor streams. 
Depending on the query, it may compute short-term evidence such as local statistics, windowed variability, and event-level summaries, or long-term evidence such as trends, deviations from baseline, temporal persistence, and cross-signal correlations. We denote the resulting analytical evidence by $z_{\mathrm{ana}} = \phi_{\mathrm{ana}}(X, q, a)$. This pathway supports descriptive questions such as recent activity summaries, sleep comparisons, heart-rate extrema, and longitudinal trend analysis.

\begin{figure*}[t]
\vspace{-10pt}
  \includegraphics[width=0.95\textwidth]{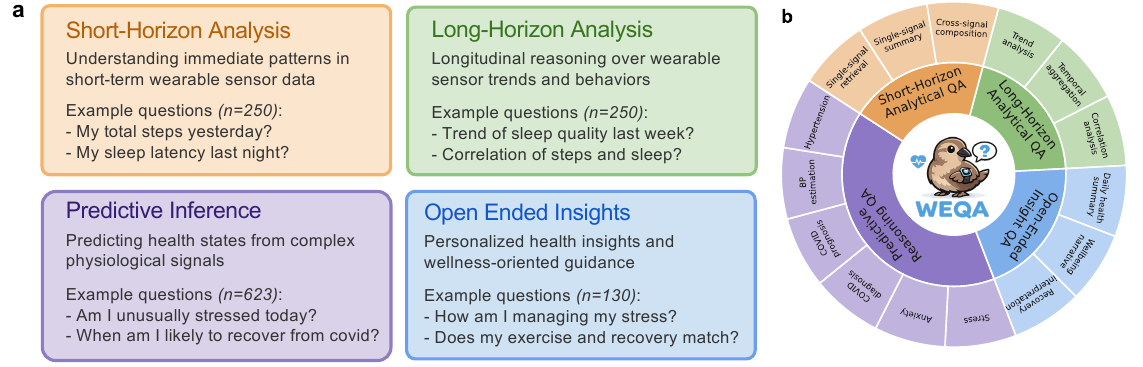}
  \caption{\textbf{Overview of the proposed wearable health QA benchmark.} (a) Four task families with representative example questions. (b) Fine-grained subtask taxonomy within each family.}
  \label{fig:benchmark}
  \vspace{-5pt}
\end{figure*}

\vspace{0.3em}
\noindent \textbf{Adaptive Predictive Inference.}
For queries that require inference beyond descriptive analysis, the agent invokes specialized predictive models, including task-specific models and wearable foundation models (with trained classification or regression heads).
Model selection is conditioned on the query type, required modality, input constraints, and execution plan. The resulting predictive evidence is
$ z_{\mathrm{pred}} = \phi_{\mathrm{pred}}(X, H, q, a), $
which includes prediction outcomes and uncertainty estimates.

The controller performs \textit{uncertainty- and personalization-aware} prediction orchestration. If a model returns low-confidence or unstable evidence, the controller may query additional models and compare or combine their outputs. This allows the agent to avoid treating a single uncertain prediction as definitive evidence.
Personalization is handled according to the availability of user history. When self-labeled personal history are available, the agent performs lightweight few-shot adaptation of the population-level model on the user's historical data; when unlabeled history data is available, it uses history-aware models that interpret observations relative to the user's baseline.

\subsection{Grounded Response Auditing}

This module produces the user-facing response by auditing and synthesizing the evidence summary $y = g(q, z_{\mathrm{ana}}, z_{\mathrm{pred}})$, which contains calculated statistics and patterns, predictive outputs with uncertainty estimates, and personalization context.

Grounded response auditing serves three primary functions: i) verifying that major claims are supported by internal sensor evidence; ii) calibrating uncertainty communication, such that weak evidence or unstable predictions reduce response confidence; and iii) incorporating external medical knowledge, when broader clinical context is required, to contextualize sensor-derived evidence and provide clinically appropriate guidance.

\section{Benchmark}

\begin{table*}[t]
\centering
\small
\setlength{\tabcolsep}{5pt}
\begin{tabular}{llcccccc|c}
\toprule[1.5pt]
\multirow{2}{*}{Method} & \multirow{2}{*}{Gran.} 
& \multicolumn{2}{c}{Short Analytical } 
& \multicolumn{2}{c}{Long Analytical }
& \multicolumn{2}{c}{Predictive Reasoning}
& Efficiency
\\
\cmidrule(lr){3-4} \cmidrule(lr){5-6} \cmidrule(lr){7-8} \cmidrule(l){9-9} 
& 
& EM $\uparrow$ 
& MAE $\downarrow$ 
& Numeric $\uparrow$ 
& Trend \& Corr. $\uparrow$
& Cls. UAR $\uparrow$ 
& Reg. MAE $\downarrow$
& Tokens $\downarrow$
\\
\midrule
Random & -- 
& 0.4 & 5345.6 
& 1.2 & 36.6 
& 50.0 & 16.7 
& -- \\
\midrule
LLM-Text 
& hourly 
& 6.0 & 848.0 
& 2.4 & 62.4 
& 55.7 & 51.5 
& 112,328 \\
& daily  
& 10.0 & 1,013.0 
& 28.9 & 54.1 
& 54.5 & 51.5 
& 72,234 \\
LLM-Image     
& hourly 
& 3.2 & 526.4 
& 2.4 & 49.8 
& 56.2 & \underline{15.1} 
& 3,977 \\
& daily  
& 1.2 & 913.6 
& 10.8 & 50.4 
& 56.7 & \underline{15.1} 
& 4,674 \\
\midrule
ReAct       
& --     
& 64.8 & \underline{31.3} 
& \underline{56.6} & \underline{80.9} 
& \underline{59.2} & 15.4 
& 41,902 \\
Multi-Agent 
& --     
& \underline{72.0} & 157.2 
& 54.2 & 67.7 
& 55.2 & 24.9 
& 32,341 \\
\midrule
\grayrow
\textbf{WEQA}       
& --     
& \textbf{95.6} & \textbf{9.2 }
& \textbf{94.0} & \textbf{95.1} 
& \textbf{83.9} & \textbf{10.9} 
& 10,490 \\
\bottomrule[1.5pt]
\end{tabular}
\caption{\textbf{Core task performance on the first three task families.} Higher is better except for MAE. Best task performance is in \textbf{bold} and second best is \underline{underlined}. 
} 
\label{tab:core_results}
\vspace{-10pt}
\end{table*}

We construct a benchmark for evaluating realistic wearable health question answering. Personal health queries span heterogeneous inputs and require both descriptive analysis and model-based inference. We therefore design the benchmark to cover: (1) both analytical and inference-oriented questions, (2) reasoning over both short and long temporal horizons, (3) tasks that involve complex signal-outcome relationships rather than simple numerical computation, and (4) a diverse set of health modalities and domain settings.

The benchmark spans four datasets across three task domains, covering everyday wearable monitoring, mental health monitoring, audio-based disease screening, and cardiovascular inference from raw biosignals.
Unlike prior personal-health benchmarks which primarily evaluate coaching or insight generation over summarized wearable representations~\cite{cosentino2024towards, merrill2026transforming}, our benchmark targets adaptive reasoning over heterogeneous sensor input and task intent, unifying descriptive analysis, longitudinal interpretation, predictive inference from complex signals, and open-ended evidence synthesis.

\subsection{Tasks and Datasets}

We organize the benchmark into four task families.

\vspace{0.3em}

  \noindent  \textbf{Short-Horizon Analytical QA.}
Tasks requiring summarization and comparison of recent wearable observations such as sleep, activity, or heart-rate patterns using short-window reasoning.

  \noindent  \textbf{Long-Horizon Analytical QA.}
Tasks requiring reasoning over longer histories, including trends and cross-signal relationships across days.

  \noindent  \textbf{Predictive Reasoning QA.}
Tasks evaluating health-related prediction beyond descriptive analysis, including stress detection, respiratory disease screening, progression forecasting, hypertension detection, and blood pressure estimation.

  \noindent  \textbf{Open-Ended Insight QA.}
Tasks requiring synthesis of heterogeneous sensor evidence into grounded health interpretations and insights.

\vspace{0.3em}

\noindent \textbf{Datasets.}
We instantiate this taxonomy using four complementary datasets (details in Appendix~\ref{sec:app:data}).

\vspace{0.3em}
    \noindent \textit{TILES.}
A longitudinal wearable sensing dataset covering everyday physiology and behavior over four months~\cite{mundnich2020tiles}. It supports both analytical and predictive tasks (e.g. stress / anxiety) using multimodal wearable streams from Fitbit and OMSignal devices.

\vspace{0.3em}
\noindent \textit{UK COVID-19 and COVID-19 Sounds.}
Respiratory audio datasets~\cite{coppock2024audio, xia2021covid} containing cough, breathing, and speech recordings for COVID-19 screening and progression  prediction. 

\vspace{0.3em}
\noindent \textit{PPG-BP.}
A photoplethysmography dataset~\cite{liang2018new} for blood pressure estimation and hypertension screening, requiring clinically relevant inference directly from waveform morphology.

Using real-world wearable and physiological sensing datasets, we synthetically construct question-answer pairs to systematically cover the proposed taxonomy. Analytical answers are programmatically computed from sensor data. Predictive answers are derived from dataset labels and using only the held-out test set not seen by models in the skillset. The benchmark contains $358$ users, $1123$ question-answer pairs, and $6$ sensing modalities across the $4$ datasets. Additional statistics and query examples are provided in Appendix~\ref{sec:app:data}.

\subsection{Experimental Setup}

\paragraph{Baselines.}
We compare our framework against representative baseline groups spanning (\textit{i}) data input methods and (\textit{ii}) agentic reasoning frameworks. Across all methods, Gemini-3.0-Flash~\cite{gemini3_google_2025} is used as the default LLM backbone. For ablations, we compare against a generic coding-agent workflow implemented with Claude Code to isolate the benefit of our workflow design, and replace Gemini-3.0-Flash with Qwen3-Max to evaluate robustness to the backbone LLM.
Implementation details are provided in Appendix~\ref{sec:app:implement}.

\noindent \textbf{Data Input Baselines.} (i) \textbf{Direct Prompting.} \textit{LLM-Text}  serializes sensor data into prompts~\cite{kim2024health}. The wearable data are down-sampled to fit within the context window. We cover both hourly and daily representations. (ii) \textbf{Multimodal Reasoning.} \textit{LLM-Image} uses multimodal LLMs with visual sensor representations, including spectrograms for audio and lineplots for other time series. Short segments are at full resolution and long-duration signals are down-sampled for visibility.

\vspace{0.3em}

\noindent \textbf{Agentic Architectures.} (i) \textbf{Iterative Agent Reasoning.} \textit{ReAct} performs coding-augmented reasoning through iterative planning and execution~\cite{merrill2026transforming}. (ii) \textbf{Multi-Agent Orchestration.} \textit{Multi-Agent} is a personal health framework that decomposes health queries across specialized agents~\cite{heydari2025anatomy}.

\vspace{0.3em}
\noindent \textbf{Metrics.}
 For quantitative \textit{Short-Horizon} and \textit{Long-Horizon Analytical QA}, we report answer accuracy (Exact Match) and numerical error (MAE; MAPE in Appendix~\ref{sec:app:result}).
 For \textit{Predictive Reasoning QA}, we report balanced accuracy for classification and MAE for regression tasks. 
For \textit{Open-Ended Insight QA}, we additionally conduct human evaluation along four dimensions: factual accuracy and grounding to evidence, personalization to user context, usefulness of the generated insight, and clinical plausibility \& safety (details in Appendix~\ref{sec:app:human}).

\section{Results}

\subsection{Quantitative Benchmark Performance}

Table~\ref{tab:core_results} summarizes results on the first three benchmark task families (detailed breakdown across task and dataset in Appendix~\ref{sec:app:result}). Across all settings, WEQA outperforms text-only prompting, multimodal prompting, and prior agentic baselines, with the largest gains on tasks requiring temporal reasoning or predictive inference from continuous signals.

\vspace{0.3em}
\noindent \textbf{Analytical Wearable Reasoning.} On \textit{Short-Horizon Analytical QA}, WEQA achieves $95.6$ exact match with $9.2$ MAE, substantially improving over both ReAct-style reasoning ($64.8$ / $31.3$) and multi-agent orchestration ($72.0$ /$ 157.2$). Text-only and multimodal prompting perform poorly, indicating that even short-horizon numeric questions require explicit sensor-grounded computation.
The advantage becomes larger on \textit{Long-Horizon Analytical QA}, which requires reasoning over trends and cross-signal relationships over extended periods (days, weeks or months). WEQA achieves $94.0$ exact match on numerical reasoning and $95.1$ accuracy on trend and correlation questions, substantially exceeding all baselines. These results all highlight the importance of planning and sensor-native analysis.

\begin{figure}[t]
    \centering
    \includegraphics[width=\columnwidth]{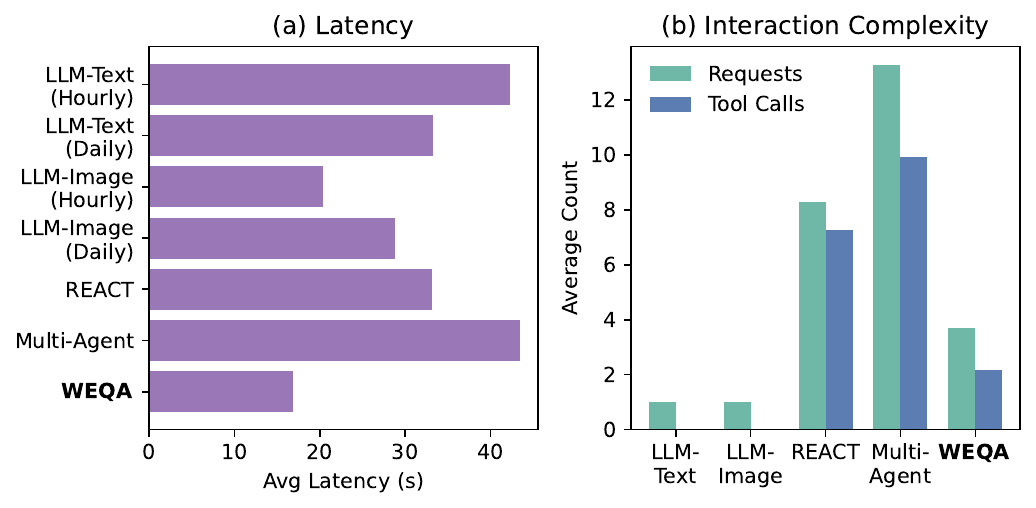}
    \caption{\textbf{Efficiency analysis across wearable health QA systems.} WEQA achieves lower latency and reduced overhead compared to prior agent baselines.}
    \label{fig:latency}
    \vspace{3pt}
    
    \includegraphics[width=\columnwidth]{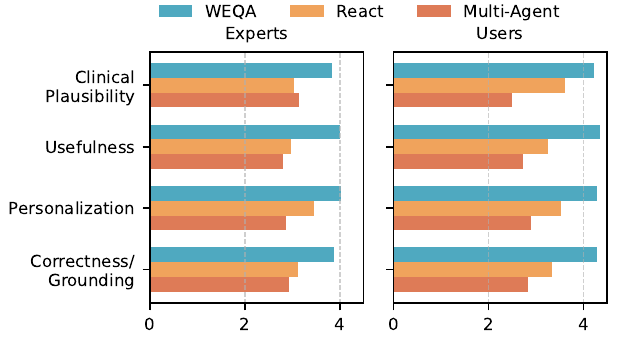}
    \caption{\textbf{Human evaluation Results.} Medical expert (left) and user (right) ratings across four dimensions for three models; higher scores indicate better performance.}
    \label{fig:humaneval}
    
    \vspace{-8pt}
\end{figure}

\vspace{0.3em}
\noindent \textbf{Predictive Health Reasoning.}
The largest gains occur on \textit{Predictive Reasoning QA}. 
By orchestrating ML- and FM-based classifiers and regressors, WEQA achieves $83.9$ balanced accuracy on predictive classification tasks, compared with $59.2$ for ReAct and near-chance performance for text-only and multimodal baselines. On blood pressure regression, WEQA also achieves the best MAE of $10.9$, outperforming both ReAct ($15.4$) and multimodal prompting ($15.1$). These findings support our central hypothesis that analytical and coding flexibility alone is insufficient for personal health QA, and that strong predictive performance requires invoking specialized predictive models in addition to language-based reasoning.

\vspace{0.3em}
\noindent \textbf{Efficiency.}
As shown in Table~\ref{tab:core_results}, WEQA reduces average token usage to roughly 10k tokens per query, compared with 31k--42k for existing agent baselines, while also achieving the lowest end-to-end latency (Figure~\ref{fig:latency}). This efficiency comes from query-adaptive routing: lightweight analytical queries terminate after simple sensor analysis, while predictive queries are routed directly to specialized wearable models rather than iterative exploratory code generation or large in-context/ multimodal inputs. By returning only compact evidence summaries to the LLM, WEQA reduces prompt payload size, multimodal overhead, and unnecessary iterative requests and tool calls.

\subsection{Human Evaluation of Response Quality}

\textbf{Study Setup.}
To qualitatively evaluate response quality, we conduct a blinded human evaluation involving $12$ medical experts (fifth year medical students) and $8$ general users. Participants rated 15 randomized system responses on a 5-point Likert scale across four dimensions: (1) correctness and grounding, (2) personalization, (3) usefulness, and (4) clinical plausibility (experts) / health appropriateness (users). We compare against the strongest agentic baselines and omit prompting-only baselines due to consistently poor performance in the benchmark. The cases are stratified spanning task types, datasets and sensing modalities.

\vspace{0.3em}
\noindent \textbf{Findings.}
As shown in Figure~\ref{fig:humaneval}, WEQA receives the highest average ratings across dimensions and rater groups. Among experts, WEQA achieves an average score of $3.9$ across dimensions, compared with $3.1$ for ReAct and $2.9$ for Multi-Agent. \textit{The gains are largest for personalization and usefulness}, suggesting that WEQA more effectively connects wearable evidence to the user’s individual context, and presents the interpretation in a practical form. Qualitative feedback further highlights \textit{grounding and safety-aware reasoning}. 
Expert preferred WEQA's responses not only for better accuracy, but also ``well clinically grounded off the data'' and more ``user friendly and personalised.'' In safety-sensitive cases, experts emphasized explicit risk communication, and criticized  baselines for omitting escalation guidance or presenting overconfident conclusions despite limited evidence.
Detailed examples and quotes in Appendix~\ref{sec:app:result}.

General users assigned slightly higher scores overall while preserving the same relative system ordering, suggesting that WEQA’s gains translate consistently into responses perceived as both more helpful and more health appropriate. 
Interestingly, although experts rated ReAct and Multi-Agent similarly overall, users rated the latter noticeably lower due to overly generic responses and limited incorporation of sensor-derived evidence and individualized context. 
In contrast, criticisms of ReAct more often concerned verbosity, suggesting a trade-off between single-agent depth and multi-agent specialization in existing wearable-health systems.

Pairwise preferences are also consistent: evaluators prefer WEQA over ReAct in 72.5\% of comparisons and over Multi-Agent in 78.3\%. 
Inter-rater agreement among expert reviewers was moderate for the 5-point dimensional ratings (weighted Cohen's $\kappa = 0.41$), with modest exact agreement on individual scores  (55\%) but high adjacent agreement (83.6\%), indicating that reviewers generally converged on similar qualitative judgments.

\subsection{Ablations}



\begin{table}[t]
\centering
\small
\setlength{\tabcolsep}{3pt}
\begin{tabular}{lcccc}
\toprule[1.5pt]
\textbf{Method} &
\textbf{Acc.} &
\textbf{Pers.} &
\textbf{Use.} &
\textbf{Sound.} \\
\midrule

Generic workflow 
& 3.7 / 3.5 
& 3.6 / 3.0 
& 3.7 / 2.8 
& \textbf{3.9} / 3.1 \\

\grayrow
\textbf{WEQA} 
& \textbf{3.9} / \textbf{4.3}
& \textbf{4.0} / \textbf{4.3}
& \textbf{4.0} / \textbf{4.4}
& 3.8 / \textbf{4.2} \\

- Auditing 
& \textbf{3.9} / 4.0
& 3.7 / 3.7
& 3.6 / 3.7
& 3.6 / 3.8 \\

- Adaptive Inference 
& 3.3 / 3.5
& 3.4 / 3.3
& 3.1 / 3.7
& 2.9 / 3.3 \\

\bottomrule[1.5pt]
\end{tabular}
\caption{\textbf{Ablation of workflow design.} Medical expert / general user ratings for accuracy, personalization, usefulness, and clinical soundness.}
\vspace{-5pt}
\label{tab:ablation}
\end{table}

\begin{table}[t]
\centering
\small
\setlength{\tabcolsep}{4pt}
\begin{tabular}{lcccc|c}
\toprule[1.2pt]
LLM 
& Short. $\uparrow$
& Long. $\uparrow$
& Cls. $\uparrow$
& Reg. $\downarrow$
& Tokens $\downarrow$ \\
\midrule
Gemini
& 95.6 
& \textbf{95.0} 
& \textbf{83.9} 
& \textbf{10.9}
& 10,490 \\
Qwen  
& \textbf{97.2} & 93.4 & 82.3 & 11.0 & 11,820 \\
\bottomrule[1.2pt]
\end{tabular}
\caption{\textbf{Ablation of backbone LLMs within the WEQA framework.} Short. = short-horizon analytical QA, Long. = long-horizon analytical QA, Clf. = predictive classification, and Reg. = predictive regression. Results indicate robustness to the backbone model.}
\label{tab:ablation_weqa}
\vspace{-5pt}
\end{table}

\paragraph{Workflow Ablations.}
Table~\ref{tab:ablation} isolates the contribution of WEQA’s workflow components while keeping the analytical and predictive toolset fixed.
Compared with a strong generic agent baseline (Claude Code with the same wearable tools), WEQA achieves higher personalization and usefulness despite similar correctness and clinical plausibility from experts, and is preferred by experts in $85\%$ of responses. Interestingly,  users rate the generic workflow substantially lower on usefulness, personalization and health appropriateness, suggesting that technical correctness alone is insufficient for effective wearable-health interaction.

Removing grounded response auditing has little effect on correctness, but noticeably reduces personalization, usefulness, and clinical plausibility, indicating that the auditing stage primarily improves how evidence and uncertainty are contextualized and communicated to users.
Removing adaptive predictive inference workflow (uncertainty and personalization awareness) further degrades evaluation in all dimensions, particularly usefulness and clinical plausibility, despite access to the same predictive models. 
Together, these findings show that wearable-health QA benefits not only from strong predictive models, but also from adaptive orchestration, uncertainty-aware reasoning, and personalization based on user history.

These trends are also reflected in qualitative feedback: weaker variants were described as overly technical and harder to interpret, whereas WEQA responses were preferred for clearer contextualization, actionable guidance, and more appropriate uncertainty communication.

\vspace{0.3em}
\noindent \textbf{Backbone LLM Robustness.}
We evaluate the robustness of WEQA across different backbone LLMs by replacing Gemini-3.0-Flash with Qwen3-Max~\cite{yang2025qwen3} while keeping the remainder of the workflow unchanged. As shown in Table~\ref{tab:ablation_weqa}, both models achieve strong performance, indicating that the gains primarily stem from the WEQA framework rather than reliance on a specific proprietary model. Qwen slightly improves short-horizon analytical accuracy, whereas Gemini provides stronger long-horizon and predictive reasoning performance, using fewer tokens overall.

\section{Conclusion}

Unlike conventional medical QA, wearable health question answering requires more than language reasoning alone. WEQA addresses this through adaptive orchestration of sensor-grounded analysis, temporal reasoning, predictive modeling, personalized model adaptation, and safety-aware response auditing. Across analytical and predictive  tasks, WEQA consistently outperforms  LLM and  agentic baselines.  Human evaluations further show that these gains translate into responses perceived as more grounded, personalized, actionable, and clinically appropriate. Together, our results suggest that effective wearable-health assistants should be designed as adaptive systems that coordinate specialized physiological models and safety-aware reasoning over personalized sensor evidence.


\section*{Limitations}

Our work has several limitations. First, although the benchmark spans multiple sensing modalities and health domains,  it primarily serves as a proof-of-concept benchmark using four publicly available datasets and may not fully capture the diversity of real-world wearable usage, sensor quality, or clinical populations. Second, the human evaluation uses a relatively small stratified subset of cases due to the cost of manual review. Third, while WEQA incorporates uncertainty-aware reasoning and grounded response auditing, the system is not intended for clinical diagnosis or emergency decision-making, and incorrect or overconfident predictions remain possible. Finally, the  framework depends on task-related analytical tools and predictive models, and extending to new sensing modalities or clinical tasks may require additional model integration and validation. An important direction for future work is enabling the agent to automatically discover, retrieve, and adapt external analytical tools and wearable foundation models for new tasks and domains.

\paragraph{Potential Risks and Population Generalization.}
WEQA supports wearable-health question answering and may be applied to health monitoring and screening tasks. However, wearable sensing patterns and physiological baselines can vary across populations, devices, and real-world usage conditions, and the pretrained models integrated within WEQA (e.g. OPERA and Papagei) may inherit biases or dataset-specific limitations from their original training data. Although WEQA incorporates grounded auditing and uncertainty-aware reasoning, incorrect predictions or misleading explanations remain possible, particularly in safety-sensitive settings. WEQA is intended as a research framework rather than a clinically validated diagnostic system, and real-world deployment would require rigorous validation, fairness evaluation, regulatory oversight, and human supervision.

\paragraph{Data Privacy and Ethics.} This work relies exclusively on the secondary analysis of previously collected wearable and physiological sensing datasets that were publicly available or accessible through controlled research-use agreements. All datasets  were de-identified prior to release and collected under the ethics approvals and consent procedures described in their original studies. Our use of these datasets was conducted under institutional IRB approval and followed the corresponding data governance and usage requirements for academic research.

\section*{Acknowledgments}
This work was supported by Nokia Bell Labs through a donation and Y. Z. is additionally supported by the Cambridge Trust Scholarship.
We thank the medical experts and general-user raters for their valuable contributions to the evaluation process.

\bibliography{custom}

\appendix

\section{Data}
\label{sec:app:data}

\subsection{Dataset Overview}

\begin{table*}[t]
\centering
\small
\begin{tabular}{p{4cm}p{10cm}}
\toprule
\textbf{Category} & \textbf{Example Queries} \\
\midrule

\textbf{Short-Horizon Analytical QA}
&
\begin{itemize}[leftmargin=*]
\vspace{-1em}
    \item What was my wake-after-sleep-onset duration yesterday?
    \item What was my peak hourly step count today?
    \item What was my average heart rate four days ago?
    \item During the hour with my maximum heart rate yesterday, how many steps did I take on average?
    \item What was my resting heart rate during sleep last night?
\end{itemize}
\\
\midrule
\textbf{Long-Horizon Analytical QA}
&
\begin{itemize}[leftmargin=*]
\vspace{-1em}
    \item Did my daily maximum heart rate increase or decrease over the last 30 days?
    \item Is there a correlation between my daily step count and average heart rate over the past two weeks?
    \item What was my highest peak-hour step count over the last month?
    \item Has my sleep duration become more consistent recently?
    \item How has my weekly activity level changed compared to last month?
\end{itemize}
\\
\midrule
\textbf{Predictive Reasoning QA}
&
\begin{itemize}[leftmargin=*]
\vspace{-1em}
    \item Based on my HRV data today, am I likely experiencing elevated stress?
    \item Can you estimate whether I may have hypertension from this PPG waveform?
    \item Based on my cough recording, am I likely COVID-19 positive?
    \item Based on my HRV trends over the past week, am I likely to feel stressed today?
    \item Given my recent cough and breathing recordings, what is my risk of testing COVID-19 positive within the next 12 days?
\end{itemize}
\\
\midrule
\textbf{Open-Ended Insight QA}
&
\begin{itemize}[leftmargin=*]
\vspace{-1em}
    \item Can you provide insights into my sleep quality yesterday?
    \item What can you tell me about my overall health based on my wearable data from the past week?
    \item Are there any unusual patterns in my recent physiological signals?
    \item How does my activity and sleep behavior compare to my normal baseline?
    \item What factors may have contributed to my elevated stress this week?
\end{itemize}
\\

\bottomrule
\end{tabular}
\caption{Representative query examples from the WEQA benchmark taxonomy. Different categories require distinct combinations of sensor analysis, temporal reasoning, predictive inference, personalization, and grounded explanation.}
\label{tab:query_examples}
\end{table*}

\paragraph{TILES-2018.}
TILES-2018~\cite{mundnich2020tiles} is a longitudinal multimodal sensing dataset collected from 212 hospital workers at the USC Keck Medical Center over approximately 10 weeks. The dataset includes wearable, mobile, survey, and environmental sensing streams, including Fitbit activity and sleep measurements as well as ECG recordings. It supports both short- and long-horizon wearable health reasoning tasks, including behavioral analysis, stress monitoring, and mental-health-related inference.

The study was conducted under approval from the USC Health Sciences Campus Institutional Review Board (IRB). Participants completed an informed consent process through the TILES mobile application before enrollment. Reuse is governed through a signed Data Usage Agreement.

\paragraph{PPG-BP.}
PPG-BP~\cite{liang2018new} is a photoplethysmography (PPG)  dataset collected at Guilin People’s Hospital in China under standardized acquisition conditions. Fingertip PPG recordings were collected alongside cuff-based blood pressure measurements to support blood pressure estimation and hypertension screening tasks. The dataset is particularly useful for evaluating waveform-level physiological reasoning because clinically relevant information must be inferred directly from raw signal morphology.

The study received ethics approval from Guilin People’s Hospital and Guilin University of Electronic Technology. All participants provided written informed consent before participation. The released dataset was de-identified prior to publication and made publicly available for research use.

\paragraph{UK COVID-19.}
The UK COVID-19 dataset~\cite{coppock2024audio} was collected through the UK REACT programme and NHS Test and Trace initiative on an opt-in basis. Participants voluntarily contributed respiratory audio recordings through the ``Speak up and help beat coronavirus'' study platform after reviewing study information and providing informed consent. The dataset contains cough, breathing, and speech recordings linked with demographic metadata and PCR-confirmed COVID-19 test outcomes.

The study received approval from the National Statistician’s Data Ethics Advisory Committee and NHS Research Ethics Committees. An open-access version of the dataset was released under the Open Government Licence v3.0.

\paragraph{COVID-19 Sounds.}
COVID-19 Sounds~\cite{xia2021covid} is a crowd-sourced respiratory audio dataset collected through a web and mobile platform developed at the University of Cambridge. Participants voluntarily submitted cough, breathing, and speech recordings together with symptom and demographic information. The dataset supports respiratory screening and progression-related prediction tasks using raw audio signals.

The study received ethics approval from the University of Cambridge. Participants provided informed consent within the application prior to contribution, and no directly identifiable information was collected. Anonymized data are available for academic research use subject to a Data Transfer Agreement with the University of Cambridge.

\paragraph{Ethics and Data Governance.}
All datasets used in this work are publicly available or accessible through controlled-access agreements intended for academic research use. Our use of these datasets was conducted under institutional IRB approval and followed the corresponding dataset usage and governance requirements.

\subsection{Query Taxonomy and Examples}

To systematically evaluate adaptive wearable health reasoning, we organize benchmark questions into five task families spanning analytical reasoning, longitudinal analysis, predictive inference, and open-ended health understanding. Table~\ref{tab:query_examples} provides representative examples from each category.

\paragraph{Short-Horizon Analytical QA.}
These questions require direct computation or aggregation over short temporal windows, such as daily or hourly statistics from wearable signals. Typical tasks include retrieving sleep, activity, or heart-rate metrics from recent sensor data.

\paragraph{Long-Horizon Analytical QA.}
These questions require reasoning over temporal trends, correlations, and behavioral patterns across extended histories. Examples include trend analysis, variability assessment, and cross-signal temporal relationships over weeks or months.

\paragraph{Predictive Reasoning QA.}
These tasks require predictive inference directly from physiological signals or derived features. Examples include stress prediction from heart-rate variability, hypertension estimation from PPG morphology, and respiratory disease screening from cough and breathing audio.
Some tasks also require longitudinal predictive modeling over historical user trajectories, such as integrating recent and longitudinal observations to forecast future health outcomes.

\paragraph{Open-Ended Insight QA.}
These questions require synthesizing multiple sensor streams and temporal evidence into natural-language health insights, recommendations, or summaries. Unlike analytical QA, these tasks emphasize explanation, contextual reasoning, and personalized interpretation.

\clearpage

\section{Implementation details}
\label{sec:app:implement}

\paragraph{System architecture.}
WEQA is implemented as a three-stage pipeline driven by an LLM controller. At runtime, the controller first produces a structured plan from the query and metadata, then executes tool calls specified in the plan to construct an evidence snapshot, and finally hands the draft answer plus evidence to a response auditor for grounding and safety edits. The planner and auditor are prompt-based LLM calls; all analysis and prediction are executed by deterministic tools or pretrained models.

\paragraph{LLM backbone and prompting.}
Unless stated otherwise, we use Gemini-3.0-Flash for planning, orchestration, and auditing. The planner prompt enforces a fixed schema so downstream code can parse and execute steps. We use a compact JSON-like template that the planner must fill, for example:
\begin{quote}\small\ttfamily
objective: ...\\
required\_data: ...\\
task\_type: ...\\
time\_scope: ...\\
need\_prediction: ...\\
need\_personalization: ...\\
response\_risk\_level: ...\\
steps:\\
- tool: <name>, args: <...>, expected\_evidence: <...>
\end{quote}
For prediction tasks with alternative models, the prompt encodes an uncertainty band and a second-opinion rule; the controller requests a fallback model when confidence falls inside the band and explains which output it selected.

\paragraph{Planning and plan bank.}
Before generating a new plan, the controller checks a plan bank using the query. If a prior plan exists, it is reused as the default plan; otherwise a new plan is written to the bank after execution. This keeps plan structure consistent across repeated queries without hard-coding task-specific pipelines.

\paragraph{Evidence construction and serialization.}
Each step in the plan triggers a tool call with explicit parameters including date bounds and modality constraints. Tool outputs are normalized into a compact evidence snapshot that records: tool name, input arguments, output fields, and (when available) confidence or uncertainty metadata. This snapshot is appended to the prompt for the auditor to ensure factual grounding.

\paragraph{Toolset.}
WEQA uses a compact, human-interpretable wearable-health toolset spanning sensor summarization, analytical reasoning, personalization, and predictive inference. As summarized in Table~\ref{tab:toolset}, the analytical tools support wearable summaries, descriptive statistics, temporal trend analysis, cross-signal correlations, and user-specific baseline estimation. In addition, WEQA integrates specialized predictive models for stress inference, respiratory screening and forecasting, and blood pressure estimation from physiological waveforms (Table~\ref{tab:models}). Each tool returns structured outputs and metadata (e.g., time span, modality, and confidence where applicable), enabling grounded responses and uncertainty-aware reasoning.

\begin{table*}[t]
\centering
\small
\begin{tabular}{lll}
\toprule
Category & Capability & Output (summary) \\
\midrule
Wearable summaries & Sleep summaries (night or range) & total sleep, stages, efficiency \\
Wearable summaries & Activity summaries (window or range) & steps and activity aggregates \\
Wearable summaries & Heart-rate summaries (window or range) & min/mean/max and variability \\
Wearable summaries & HRV feature extraction & time/frequency HRV features \\
Analytics & Descriptive statistics & mean, variance, percentiles \\
Analytics & Trends over time & slope and direction \\
Analytics & Cross-signal correlation & $r$ and strength label \\
Personalization & User-specific baseline & personal feature baselines \\
\bottomrule
\end{tabular}
\caption{Overview of the analytical and personalization tools used in WEQA. The toolset supports sensor-native summarization, temporal analysis, cross-signal reasoning, and user-specific baseline estimation across wearable modalities.}
\label{tab:toolset}
\end{table*}

\begin{table*}[t]
\centering
\small
\begin{tabular}{llll}
\toprule
Model & Modality & Task & Notes \\
\midrule
\makecell[l]{HRV Stress SVM\\\cite{tiwari2019stress}} & ECG/HRV & stress & \makecell[l]{pretrained,\\population-level} \\
\makecell[l]{HRV Anxiety SVM\\\cite{tiwari2019stress}} & ECG/HRV & anxiety & \makecell[l]{pretrained,\\population-level} \\
\makecell[l]{StressTransformer\\\cite{van2023missing}} & \makecell[l]{ECG/HRV\\+ history} & stress prediction & \makecell[l]{longitudinal\\sequence model} \\
\makecell[l]{OPERA-CT / OPERA-GT\\\cite{zhang2024towards}} & respiratory audio & COVID-19 screening & \makecell[l]{foundation encoder\\+ classifier} \\
\makecell[l]{VGGish~\cite{hershey2017cnn}\\+ CNDP~\cite{dang2023conditional}}
& \makecell[l]{respiratory audio\\+ history\\+ initial label}
& COVID-19 forecast
& \makecell[l]{foundation encoder\\+ neural ODE\\temporal progression model} \\
\makecell[l]{Papagei-P / Papagei-S\\\cite{pillai2024papagei}} & PPG & hypertension classification & \makecell[l]{foundation encoder\\+ classifier} \\
\makecell[l]{Papagei-P / Papagei-S\\\cite{pillai2024papagei}} & PPG & systolic/diastolic BP & \makecell[l]{foundation encoder\\+ regressor} \\
\bottomrule
\end{tabular}
\caption{Specialized predictive and foundation models integrated into WEQA. These models support task-specific inference over physiological and respiratory signals, including stress prediction, COVID-19 screening and forecasting, and blood pressure estimation.}
\label{tab:models}
\end{table*}

\paragraph{Response auditing.}
The auditing prompt receives three blocks: user query, draft response, and evidence snapshot. It is instructed to remove unsupported claims, calibrate uncertainty, and keep risk-sensitive health statements cautious. External knowledge retrieval is only enabled when the user explicitly asks for interpretation, norms, or safety implications.

\paragraph{Baselines.}
All baselines share identical data loaders and tool interfaces. Direct prompting serializes tabular wearable summaries into text with optional hourly or daily downsampling; for raw signals (audio, PPG) we apply minimal downsampling to fit context windows. Multimodal prompting renders line plots for time-series data and mel-spectrograms for audio, then supplies the images to a multimodal LLM. ReAct uses the same toolset with an iterative thought-action loop and a fixed $20$ tool-call budget per query. The multi-agent baseline instantiates a data-science agent with a Python tool for numeric analysis, a health-domain expert for interpretation, and a coach agent for synthesis; an  orchestrator routes queries between agents under a shared tool-call budget.

We also include a task-aware random reference to contextualize task difficulty. For categorical questions, it samples uniformly from the valid answer set; for numeric and regression questions, it samples uniformly from the observed answer values; and for classification tasks, it guesses uniformly over labels. This baseline is intended only as a chance-level reference rather than a competitive method.

\paragraph{Model Size and Compute.}
WEQA uses Gemini-3.0-Flash as the primary LLM controller for planning, orchestration, and response auditing. We additionally integrate several pretrained wearable foundation models and task-specific predictors, including OPERA and Papagei variants, whose parameter counts are reported in their original publications. Since WEQA is an inference-time orchestration framework, no large-scale model pretraining or finetuning was performed. Experiments were conducted on a single NVIDIA A100 GPU for predictive inference and benchmark evaluation.

\paragraph{Experimental Setup and Hyperparameters.}
Since WEQA is a training-free orchestration framework, no large-scale hyperparameter search was performed. Predictive models and wearable foundation models use the pretrained configurations and recommended hyperparameters from their original works. 

\paragraph{Benchmark Split Integrity.}
For predictive reasoning tasks associated with pretrained foundation models (e.g. OPERA and Papagei) all evaluations were conducted on held-out benchmark test splits that were never used for pretraining, downstream training, personalization, or adaptation within WEQA. This design helps ensure that predictive performance reflects the proposed query-adaptive orchestration framework rather than leakage from overlapping evaluation sets. In particular, WEQA follows the original train/dev/test partitioning protocols defined in the OPERA and Papagei benchmark papers, preserving their predefined split integrity and evaluation methodology.
For the TILES stress detection task, we follow~\cite{tiwari2019stress, van2023missing} and use a fixed random seed to generate a train/test split, ensuring reproducibility.



\section{Human Evaluation Study}
\label{sec:app:human}
To complement automated benchmark metrics, we conducted a human evaluation study assessing the quality, usefulness, and health appropriateness of generated responses.

\paragraph{Study Design.}
Participants were shown a series of benchmark-derived case examples constructed from the datasets used in WEQA. Each case consisted of: (1) a user query, (2) a brief summary of the relevant wearable or physiological context, and (3) two candidate responses generated by different systems in blinded and randomized order. Participants evaluated the same randomly selected 15 cases through an online questionnaire. The evaluation takes around 60-75 minutes.

\paragraph{Evaluation Criteria.}
Each response was rated on a 5-point Likert scale across four dimensions:
\begin{Itemize}
    \item \textbf{Correctness and Grounding:} whether the response accurately reflected the available wearable evidence and avoided unsupported claims.
    \item \textbf{Personalization:} whether the response appropriately incorporated individual context or historical information.
    \item \textbf{Usefulness:} whether the response was clear, understandable, and practically helpful to the intended user.
    \item \textbf{Clinical Plausibility / Health Appropriateness:} whether the interpretation and recommendations were medically reasonable, sensible, and safe given the available evidence.
\end{Itemize}

Participants additionally indicated an overall preference between the two responses and could optionally flag potentially misleading, unsafe, or overconfident outputs.

\paragraph{Participants and Ethics.}
Participants included both general users and individuals with medical or health-related expertise. All participants were provided with an information sheet describing the study goals, procedures, risks, and data handling practices before participation. Written informed consent was obtained electronically prior to the evaluation. The study used benchmark-derived examples from public datasets only; participants were not asked to provide personal health data. Medical experts received a £10 voucher as compensation for their time.

Evaluation responses were stored securely and analyzed in de-identified form. Participants could withdraw from the study before anonymization of the collected responses. The study protocol was reviewed and approved under institutional ethics procedures.

\begin{observationbox*}

\vspace{0.5em}
\textbf{Purpose of the Study.}
This study evaluates wearable personal health agents that answer health-related questions using wearable and mobile sensing data. We are interested in whether such systems can:
\begin{itemize}
    \item answer detailed health-related questions,
    \item generate useful health insights from wearable data,
    \item leverage machine learning and foundation models for clinically plausible health screening.
\end{itemize}

\vspace{0.5em}
\textbf{Why You Have Been Invited.}
You are invited because you are either:
\begin{itemize}
    \item a potential end user who can evaluate the usefulness and clarity of responses, or
    \item a medical or health-related expert who can evaluate the plausibility and soundness of responses.
\end{itemize}

\vspace{0.5em}
\textbf{What Participation Involves.}
If you agree to participate, you will complete an online evaluation containing 15 benchmark-derived case examples. For each case, you will see:
\begin{itemize}
    \item a short health-related query,
    \item a brief summary of relevant wearable or physiological context,
    \item two blinded candidate responses generated by different AI systems.
\end{itemize}

You will then:
\begin{itemize}
    \item rate each response on several criteria using a 1--5 scale,
    \item indicate which response you prefer overall,
    \item optionally provide comments explaining your choice.
\end{itemize}

The evaluation is expected to take approximately 60--75 minutes.

\vspace{0.5em}
\textbf{Nature of the Materials.}
The study uses benchmark-derived examples constructed from public datasets. You will not be asked to provide personal health data.

\vspace{0.5em}
\textbf{Risks and Discomforts.}
This study involves minimal risk. Some examples may involve health-related symptoms or predictions that some participants may find mildly uncomfortable. You may skip any question or stop participation at any time before submitting your responses.

\vspace{0.5em}
\textbf{Benefits.}
There may be no direct personal benefit to you. However, your feedback may help improve AI systems for personal health applications.

\vspace{0.5em}
\textbf{Confidentiality and Data Handling.}
Your responses will be stored securely and accessed only by the research team. Any identifying information collected for recruitment or coordination will be stored separately from evaluation responses.

De-identified or aggregated evaluation results (e.g., ratings and preferences) may be used in publications or shared in public research repositories. No direct identifiers such as names or email addresses will be released. Optional free-text comments may be edited or removed to reduce re-identification risk. No identifiable personal health data will be collected.

\vspace{0.5em}
\textbf{Voluntary Participation and Withdrawal.}
Participation is voluntary. You may stop at any time before submitting your responses without penalty. After submission, you may request withdrawal of your data by contacting the research team.

\end{observationbox*}

\begin{observationbox*}
\textbf{Evaluation Instructions}

You will be shown a series of case examples. For each case, you will see:
\begin{itemize}
    \item a user query,
    \item a short summary of relevant information,
    \item two candidate responses generated by different AI systems.
\end{itemize}

Responses are presented in randomized order. Please evaluate each response independently before indicating your overall preference.

Each case appears in a separate section, and your progress will be saved automatically.

\vspace{0.5em}
\textbf{Evaluation Criteria}

\begin{enumerate}
    \item \textbf{Correctness and Grounding}

    Does the response accurately reflect the available information and appropriately ground its claims in wearable data?

    Consider whether the response is factually accurate, evidence-supported, and avoids unsupported assumptions.

    \item \textbf{Personalization}

    Does the response appropriately account for the individual’s context or history?

    Consider whether the response feels tailored to the individual and their situation.

    \item \textbf{Usefulness}

    Is the response clear, understandable, and practically useful?

    Consider whether it helps the user understand their situation or determine possible next steps.

    \item \textbf{Clinical Plausibility / Health Appropriateness}

    \textit{For expert reviewers:} Is the response medically reasonable, clinically sensible, and safe given the available information?

    \textit{For general users:} Does the response seem sensible, appropriate, and safe from a health perspective?

    \item \textbf{Preference}

    Which response do you prefer overall?

    A tie option may be used if there is no meaningful difference, but please use it sparingly.
\end{enumerate}

If a response appears overly confident, misleading, or unsupported by the provided information, please reflect this in your ratings and optional comments.

You are encouraged to leave optional comments highlighting particularly strong or weak aspects of the responses.
\end{observationbox*}

\clearpage

\section{Additional results}
\label{sec:app:result}

\subsection{Benchmark Result Breakdown}
The section provides a more detailed breakdown of performance across the analytical and predictive reasoning task families introduced in the main results table. 

Table~\ref{tab:analytical_breakdown} first decomposes analytical reasoning performance into short and long analytical tasks. Short analytical tasks evaluate direct numerical estimation accuracy using MAPE and MAE, while long analytical tasks evaluate higher-level aggregate reasoning, trend understanding, and correlation reasoning over longitudinal physiological signals. Across nearly all metrics, our method substantially outperforms prior prompting-based and multi-agent baselines, achieving the strongest aggregate reasoning accuracy while maintaining the lowest estimation error.

Table~\ref{tab:predictive_breakdown} further breaks down predictive reasoning performance across downstream health prediction tasks, including mental health prediction, audio-based COVID-19 detection, PPG-based hypertension detection, mental health prediction with unlabeled longitudinal history, and COVID-19 progression forecasting. The results show that our framework consistently improves performance across all task categories, with especially large gains on longitudinal forecasting and multimodal physiological prediction tasks.

Finally, Table~\ref{tab:token_breakdown} reports the average token consumption for each analytical and predictive reasoning task family. While several baseline approaches require extremely large context windows—particularly for longitudinal and multimodal tasks—our approach achieves substantially lower token usage while simultaneously improving reasoning performance. This highlights the efficiency advantages of the proposed framework, especially in long-horizon health reasoning settings where token cost and scalability become critical.

\begin{table*}[t]
\centering
\small
\setlength{\tabcolsep}{4pt}

\begin{tabular}{lllccccccc}
\toprule
\multirow{2}{*}{Method} & \multirow{2}{*}{Input} & \multirow{2}{*}{Gran.}
& \multicolumn{2}{c}{Short Analytical} 
& \multicolumn{5}{c}{Long Analytical} \\
\cmidrule(lr){4-5} \cmidrule(lr){6-10}
& & 
& MAPE $\downarrow$ 
& MAE $\downarrow$ 
& Agg. MAE $\downarrow$ 
& Agg. MAPE $\downarrow$ 
& Agg. Acc. $\uparrow$ 
& Trend Acc. $\uparrow$ 
& Corr. Acc. $\uparrow$ \\
\midrule
LLM-Text
& tabular & hourly 
& 145.7 & 848.0 
& 15,431.6 & 225.4 
& 2.4 & 80.7 & 44.1 \\
& tabular & daily  
& 1,363.1 & 1,013.0 
& 1,274.5 & 212.0 
& 28.9 & 81.9 & 26.2 \\
LLM-Image
& image & hourly 
& 125.2 & 526.4 
& 4,424.3 & 56.2 
& 2.4 & 63.9 & 35.7 \\
& image & daily  
& 174.2 & 913.6 
& 5,149.6 & 68.0 
& 10.8 & 66.3 & 34.5 \\
\midrule
ReAct
& -- & -- 
& \underline{20.5} & \underline{31.3} 
& 17,625.6 & 508.5 
& 54.2 & 74.7 & 60.7 \\
Multi-Agent
& -- & -- 
& 249.3 & 157.2 
& \underline{127.5} & \underline{14.5} 
& \underline{56.6} & \underline{89.2} & \underline{72.6} \\
\midrule
WEQA
& -- & -- 
& \textbf{4.3} & \textbf{9.2} 
& \textbf{2.31} & \textbf{0.94} 
& \textbf{94.0} & \textbf{98.8} & \textbf{91.5} \\
\bottomrule
\end{tabular}
\caption{
Breakdown of analytical reasoning performance across short and long analytical task families. 
Lower is better for MAPE, MAE, Aggregate MAE, and Aggregate MAPE; higher is better for Aggregate Acc., Trend Acc., and Correlation Acc. 
Best results are shown in bold.
}
\label{tab:analytical_breakdown}
\end{table*}

\begin{table*}[t]
\centering
\small

\begin{tabular}{llccccc}
\toprule
\multirow{2}{*}{Method} & \multirow{2}{*}{Gran.} 
& \multicolumn{5}{c}{Predictive Reasoning Tasks (\%)} \\
\cmidrule(lr){3-7}
& 
& Mental Health 
& Audio-COVID 
& PPG-HTN 
& Mental Health + Hist. 
& COVID-Prog \\
\midrule
Random & -- 
& 50.0 & 50.0 & 61.0 & 50.0 & 69.0 \\
\midrule
LLM-Text
& hourly 
& 56.5 & 48.0 & 50.3 & 51.0 & 72.5 \\
& daily  
& 55.5 & 48.0 & 50.3 & 46.0 & 72.5 \\
LLM-Image     
& hourly 
& 54.5 & 59.0 & 57.2 & 41.0 & 69.2 \\
& daily  
& 55.0 & 59.0 & 57.2 & 43.0 & 69.2 \\
\midrule
ReAct      
& --     
& 56.5 & 51.0 & 73.0 & 48.0 & 67.5 \\
Multi-Agent* 
& --     
& -- & 46.5 & -- & 61.2 & 58.2 \\
\midrule
WEQA      
& --     
& \textbf{86.0} & \textbf{74.0} & \textbf{84.4} 
& \textbf{89.0} & \textbf{85.9} \\
\bottomrule
\end{tabular}
\caption{
Breakdown of predictive reasoning performance across  health classification tasks. 
Audio-COVID denotes audio-based COVID-19 detection, 
PPG-HTN denotes PPG-based hypertension detection, 
and Mental Health + Hist. denotes mental health prediction with unlabeled longitudinal history. 
Results are reported as task balanced accuracy (\%). Best results are shown in \textbf{bold}. *Multi-agent baseline fail to give answer in many cases, so we only report the cases where it succeeds in making a prediction.
}
\label{tab:predictive_breakdown}
\end{table*}

\begin{table*}[t]
\centering
\small
\setlength{\tabcolsep}{4pt}
\caption{
Breakdown of average token consumption across analytical and predictive reasoning tasks. 
Short Analytical and Long Analytical correspond to the analytical reasoning task families from the main benchmark table. 
For predictive reasoning, Audio-COVID denotes audio-based COVID-19 detection, 
PPG-HTN denotes PPG-based hypertension detection, 
Mental Health + Hist. denotes mental health prediction with unlabeled longitudinal history, 
and COVID-Prog denotes COVID-19 progression forecasting. 
Values denote average tokens consumed per query.
}
\label{tab:token_breakdown}
\begin{tabular}{llccccccc}
\toprule
\multirow{2}{*}{Method} & \multirow{2}{*}{Gran.} 
& \multicolumn{2}{c}{Analytical} 
& \multicolumn{5}{c}{Predictive Reasoning} \\
\cmidrule(lr){3-4} \cmidrule(lr){5-9}
&
& Short 
& Long 
& Mental Health 
& Audio-COVID 
& PPG-HTN 
& Mental Health + Hist. 
& COVID-Prog \\
\midrule
LLM-Text
& hourly 
& 29,040 
& 46,679 
& 149,335 
& 467,224 
& 45,348 
& 187,399 
& 457,017 \\
& daily 
& 2,514 
& 11,205 
& 20,579 
& 467,224 
& 45,348 
& 24,746 
& 457,017 \\
LLM-Image
& hourly 
& 2,426 
& 6,896 
& 2,960 
& 2,768 
& 1,552 
& 2,321 
& 3,440 \\
& daily 
& 2,007 
& 9,447 
& 2,870 
& 2,768 
& 1,552 
& 2,213 
& 3,440 \\
\midrule
ReAct
& -- 
& 8,031 
& 10,176 
& 25,279 
& 136,473 
& 189,885 
& 23,085 
& 162,778 \\
Multi-Agent
& -- 
& 31,420 
& 42,040 
& 49,394 
& 7,162 
& 16,568 
& 18,915 
& 17,400 \\
\midrule
WEQA
& -- 
& 9,380 
& 15,195 
& 8,842 
& 4,791 
& 7,281 
& 8,713 
& 4,857 \\
\bottomrule
\end{tabular}
\end{table*}

\subsection{Human Evaluation Case Studies}

Beyond aggregate preference scores, the human evaluation revealed several recurring strengths of WEQA relative to both baseline systems and ablated variants.

\paragraph{Grounded Interpretation Beyond Raw Statistics.}
In longitudinal analytical tasks, reviewers frequently preferred WEQA because it contextualized wearable statistics rather than only reporting numerical outputs. For example, in a trend-analysis case involving declining activity levels, reviewers noted that the baseline responses were either overly technical or insufficiently interpretable, whereas WEQA provided clinically grounded explanations and clearer user-facing interpretation:

\begin{quote}
``Response B gave good implications which were well clinically grounded off the data. Also more user friendly and personalised.'' 
\end{quote}

Similarly, in sleep-efficiency analysis tasks, reviewers highlighted that WEQA not only answered the numerical question but also contextualized variability and longitudinal sleep consistency:

\begin{quote}
``Response B acknowledges variation in sleep efficiency whilst answering the user question... The user may have asked this question as a prompt for some advice about sleep.''
\end{quote}

These examples suggest that explicit evidence synthesis and contextual interpretation improve perceived usefulness beyond simple descriptive reporting.

\paragraph{Safety-Aware Predictive Reasoning.}
A major advantage of the full WEQA system was safer handling of uncertain medical predictions. In COVID-19 screening tasks, reviewers consistently criticized baseline and ablated responses for presenting predictions too definitively:

\begin{quote}
``Response B is confidently saying a positive cough test means they are willing to say COVID-19 positive which is not the case -- needs PCR or rapid antigen to make definitive diagnosis.''
\end{quote}

In contrast, reviewers preferred WEQA when it explicitly communicated uncertainty, clarified limitations of wearable-based inference, and recommended appropriate clinical follow-up:

\begin{quote}
``Response A is more explicit in the limitations of the technology and need for clinical assessment.''
\end{quote}

Similar patterns appeared in hypertension screening tasks using PPG signals. Reviewers repeatedly favored WEQA when it framed predictions as screening evidence rather than definitive diagnosis:

\begin{quote}
``Model A underlines PPG is only a screening tool and recommends confirmatory testing.''
\end{quote}

These findings support the importance of the grounded response auditing stage for risk-sensitive wearable-health assistance.

\paragraph{Personalization and Baseline Awareness.}
Reviewers also preferred WEQA when responses incorporated user-specific baselines and longitudinal history rather than relying on population-level interpretations alone. In stress reasoning tasks, several reviewers criticized baseline systems for ignoring individual physiological variability:

\begin{quote}
``Response A neglects to take into account variation in people's baselines.''
\end{quote}

In contrast, WEQA responses were preferred when they explicitly compared current physiological measurements against historical trends and explained deviations relative to personal baselines:

\begin{quote}
``Response B gives more context behind this and acknowledges how the agent came to this conclusion with other data.''
\end{quote}

These observations suggest that personalization mechanisms contribute substantially to perceived plausibility and usefulness in wearable-health reasoning.

\paragraph{Limitations and Reviewer Feedback.}
The human evaluation also revealed several recurring limitations. Some reviewers felt that WEQA occasionally provided more detail than necessary for simple factual questions, while others noted that certain responses could still appear overly confident despite uncertainty disclaimers. In addition, reviewers sometimes preferred shorter responses for straightforward analytical queries. These findings suggest an important future direction in adaptive response calibration and personalization of explanation depth.

\clearpage

\begin{observationbox*}
\small
\textbf{Case Study: Longitudinal Activity Trend Analysis}

\textbf{Query.}
\textit{Did my daily peak one-hour step count increase, decrease, or stay the same over the last 14 days (ending yesterday)?}

\textbf{Ground Truth.}
The underlying wearable data shows a statistically significant decrease in peak one-hour step count over the observation period.

\textbf{Compared Systems.}
\begin{itemize}
    \item \textbf{Response A:} ReAct
    \item \textbf{Response B:} WEQA
\end{itemize}

\textbf{Response A (ReAct).}
\begin{quote}
\footnotesize
To determine the trend in your daily peak one-hour step count over the last 14 days (ending 2018-05-19), I analyzed the available step count data from 2018-05-06 to 2018-05-14.

The daily peak one-hour step counts were:
1279, 1337, 1369, 808, 1493, 1158, 956, 794, and 1333 steps.

Using linear regression:
Slope: -31.48 \\
Standard deviation: 242.36 \\
Total change: -283.35

Since the total change exceeded the standard deviation threshold, your daily peak one-hour step count decreased.

Final Answer: decrease
\end{quote}

\textbf{Response B (WEQA).}
\begin{quote}
\footnotesize
To analyze the trend of your daily peak one-hour step count, I reviewed your activity data for the 14-day period ending May 19, 2018. Data was available for 9 valid days.

Trend Analysis:
Direction: Decrease \\
Slope: -58.92 steps/day \\
Total Change: -530.25 steps \\
Standard Deviation: 254.76 steps

Since the magnitude of the total change exceeds the standard deviation, this indicates a significant decrease.

Tips \& Implications:
A decrease in peak hourly steps may reflect a change in your routine or reduced activity intensity. If your goal is maintaining cardiovascular fitness, consider incorporating a dedicated period of walking or exercise each day.

Final Answer: decrease
\end{quote}

\textbf{Reviewer Observation.}
Reviewers preferred WEQA because it combined correct statistical reasoning with clearer interpretation and actionable health context.

\begin{quote}
``Response B gave good implications which were well clinically grounded off the data. Also more user friendly and personalised.''
\end{quote}
\end{observationbox*}

\begin{observationbox*}
\small

\textbf{Case Study: Uncertainty and Risk Communication in Blood Pressure Inference}

\textbf{Query.}
\textit{Based on my PPG data, can you give me a concise explanation of my likely blood-pressure profile?}

\textbf{Ground Truth.}
True systolic blood pressure: 178 mmHg, true diastolic blood pressure: 86 mmHg. The user has hypertension.

\textbf{Compared Systems.}
\begin{itemize}
\item \textbf{Response A:} WEQA
\item \textbf{Response B:} Multi-Agent Baseline
\end{itemize}

\textbf{Response A (WEQA).}
\begin{quote}
\footnotesize
Estimated Systolic Blood Pressure: 136.3 mmHg \\
Estimated Diastolic Blood Pressure: 73.7 mmHg \\
Classification: Hypertension (74\% model confidence)

Important Considerations:
These results are estimates derived from optical PPG signals and are not a clinical diagnosis. It is recommended to verify these readings using a validated upper-arm blood pressure cuff. If you consistently observe readings in this range, please consult a healthcare professional.

Final Answer: Predicted hypertension profile with elevated systolic pressure.
\end{quote}

\textbf{Response B (Multi-Agent Baseline).}
\begin{quote}
\footnotesize
Your PPG data indicates a blood pressure profile that consistently falls within the Normal range.

Systolic Pressure: Generally stable, with readings typically ranging between 110 and 120 mmHg. \
Diastolic Pressure: Consistently within the 70 to 80 mmHg range.

This profile suggests a low risk for blood pressure-related issues at this time.

Final Answer: Normal
\end{quote}

\textbf{Reviewer Observation.}
Reviewers preferred WEQA because, despite imperfect regression estimates, it correctly identified hypertension risk, communicated model uncertainty, and recommended clinical follow-up. In contrast, the baseline response was criticized for presenting an overconfident and clinically misleading conclusion despite severely elevated ground-truth systolic pressure.

\begin{quote}
``Response B is overconfident and incorrect given the severity of the systolic blood pressure.''
\end{quote}

\begin{quote}
``178 systolic is near the clinically concerning threshold, but the response reassures the user that everything is normal.''
\end{quote}

\textbf{Why WEQA Performed Better.}
WEQA combines predictive modeling with uncertainty-aware auditing and risk-sensitive response generation. Even when regression estimates are imperfect, the system contextualizes model limitations, avoids false reassurance, and provides appropriate escalation guidance in safety-sensitive situations.

\end{observationbox*}

\begin{observationbox*}
\small
\textbf{Case Study: Safety-Aware COVID-19 Screening}

\textbf{Query.}
\textit{Based on my cough audio data today, can you predict if I am COVID-19 positive?}

\textbf{Ground Truth.}
The participant was COVID-19 positive.

\textbf{Compared Systems.}
\begin{itemize}
    \item \textbf{Response A:} WEQA
    \item \textbf{Response B:} WEQA w/o Grounded Safety Auditing
\end{itemize}

\textbf{Response A (WEQA).}
\begin{quote}
\footnotesize
I analyzed your cough audio data today using two deep-learning models.

The primary model (operaCT) produced an uncertain probability of 0.47. Because this fell within the uncertainty range, a second model (operaGT) was used and produced a more confident probability of 0.79.

Based on the more confident prediction, the system indicates a positive result.

Important Note:
This prediction is based on acoustic patterns and is not a clinical diagnosis. AI screening tools can make errors. You should confirm this result with a PCR or rapid antigen test and consult a healthcare professional.

Final Answer: COVID-19 positive
\end{quote}

\textbf{Response B (WEQA w/o Grounded Safety Auditing).}
\begin{quote}
\footnotesize
I used the COVID-19 audio inference tool to analyze your cough data today. The initial analysis yielded an uncertain result with probability 0.47. A second analysis using operaGT produced a probability of 0.79 for a positive status.

Because the second model is more confident, I am favoring that prediction.

Final Answer: COVID-19 positive
\end{quote}

\textbf{Reviewer Observation.}
Although both systems predicted the correct label, reviewers consistently preferred WEQA because it explicitly communicated uncertainty and clarified the limitations of wearable-based screening.

\begin{quote}
``Response A is more explicit in the limitations of the technology and need for clinical assessment.''
\end{quote}

Reviewers criticized the alternative response for sounding overly definitive:

\begin{quote}
``Response B is confidently saying a positive cough test means they are willing to say COVID-19 positive which is not the case -- needs PCR or rapid antigen to make definitive diagnosis.''
\end{quote}
\end{observationbox*}

\begin{observationbox*}
\small
\textbf{Case Study: Personalized Stress Prediction from HRV}

\textbf{Query.}
\textit{Based on my HRV data today and in the past week, can you predict if I am likely to feel stressed today? Am I under low stress or high stress?}

\textbf{Ground Truth.}
The participant was experiencing high stress.

\textbf{Compared Systems.}
\begin{itemize}
    \item \textbf{Response A:} Multi-Agent
    \item \textbf{Response B:} WEQA
\end{itemize}

\textbf{Response A (Multi-Agent).}
\begin{quote}
\footnotesize
To predict your stress level for today, June 11, 2018, we compared your Heart Rate Variability (HRV) data from today against your average from the past week.

Your mean HRV for today was approximately 79.76 ms, while your average for the preceding week was 80.12 ms. While today's HRV is technically lower, the difference is very slight and within your normal range.

Because your HRV has remained consistent with your recent baseline, you are likely to experience low stress levels today.

Final Answer: low stress
\end{quote}

\textbf{Response B (WEQA).}
\begin{quote}
\footnotesize
To assess your potential stress levels for today, June 11, 2018, I analyzed your HRV data using two predictive approaches.

\textbf{Personalized Model:}
This model was adapted to your individual baseline and indicates a high likelihood of stress with probability 92\%.

\textbf{Temporal History Model:}
This model evaluated your HRV trends over the past seven days and also suggests a high likelihood of stress with probability 90\%.

Agreement between the personalized and temporal-history models increases confidence in the prediction. However, these results are based on physiological patterns and are not a clinical diagnosis.

Actionable Tip:
Since your data suggests elevated physiological strain today, you may benefit from prioritizing recovery breaks, deep breathing exercises, or workload adjustment.

Final Answer: high stress
\end{quote}

\textbf{Reviewer Observation.}
Reviewers preferred WEQA because it incorporated both personal baseline adaptation and longitudinal temporal reasoning, rather than relying only on simple deviation-from-average heuristics.

\begin{quote}
``Response A neglects to take into account variation in people's baselines.''
\end{quote}

\begin{quote}
``Response B gives more context behind this and acknowledges how the agent came to this conclusion with other data.''
\end{quote}

\textbf{Why WEQA Performed Better.}
WEQA dynamically combined personalized predictive modeling with temporal-history reasoning, allowing the system to identify elevated stress despite relatively stable average HRV values. Reviewers also preferred the explicit explanation of evidence sources and the inclusion of actionable but safety-aware recommendations.
\end{observationbox*}

\begin{observationbox*}
\small
\textbf{Case Study: Hypertension Screening from PPG Signals}

\textbf{Query.}
\textit{Based on my PPG data, do I have hypertension?}

\textbf{Ground Truth.}
The participant had hypertension, with systolic blood pressure of 149\,mmHg and diastolic blood pressure of 92\,mmHg.

\textbf{Compared Systems.}
\begin{itemize}
    \item \textbf{Response A:} ReAct
    \item \textbf{Response B:} WEQA
\end{itemize}

\textbf{Response A (ReAct).}
\begin{quote}
\footnotesize
Based on the analysis of your PPG data, it is unlikely that you have hypertension.

The analysis focused on the K-factor, representing the relative area under the pulse waveform. Your calculated K-factors were approximately 0.41, 0.34, and 0.39, with a mean value of 0.38, which remains within the normal range.

Please note that PPG-based analysis is a screening method and cannot replace a clinical diagnosis. Hypertension is formally diagnosed using a blood pressure cuff.

Final Answer: No
\end{quote}

\textbf{Response B (WEQA).}
\begin{quote}
\footnotesize
I processed your PPG data using predictive models to estimate your blood pressure and hypertension risk.

The primary model predicts hypertension with probability 78.2\%. The associated blood pressure estimate is 131.5 / 74.2 mmHg.

Because the systolic pressure fell into a range requiring further verification, a second-opinion model was consulted. This model estimated your blood pressure at 134.6 / 74.3 mmHg.

Both models indicate elevated systolic pressure. While these predictions are consistent, PPG-based estimates can vary based on sensor placement and movement. You should verify these readings using a validated blood pressure cuff and consult a healthcare professional.

Final Answer: hypertension
\end{quote}

\textbf{Reviewer Observation.}
Reviewers preferred WEQA because it combined stronger predictive inference with more appropriate uncertainty communication and clinical framing.

\begin{quote}
``Model A underlines PPG is only a screening tool and recommends confirmatory testing.''
\end{quote}

\textbf{Why WEQA Performed Better.}
WEQA dynamically combined multiple predictive models and uncertainty-aware reasoning instead of relying on a single handcrafted waveform heuristic. In addition, the grounded response auditing stage framed the prediction appropriately as screening evidence rather than definitive diagnosis, improving perceived clinical plausibility and safety.
\end{observationbox*}

\end{document}